\documentstyle[12pt,named]{article}
\newcommand{\qdrs}[3]
{Q$_{\mbox{\tiny #1}}$:\drs{#2}{#3}}

\newcommand{\drs}[2]
{
    {
    \begin{tabular}{|l|}
    \hline
     #1
    \\
    \hline
    #2
    \\
    \hline
    \end{tabular}
    }
}

        \catcode`@=11
        \def\*{$^{\ast}$}
        \def\?{$^{?}$}
        
        \newcounter{exai}
        \newcounter{exaii}[exai]
        \newcounter{exaiii}[exaii]
        \newcount\@exadepth \@exadepth=0
        \newenvironment{examples}%
                       {\ifnum \@exadepth >2 \@toodeep
                        \else \ifnum \@exadepth=0
                              \par
                              \begin{list}{}%
                                          {\labelwidth 3.5em%
                                           \labelsep   0.5em%
                                           \leftmargin 4.1em%
                                          }%
                              \sloppy%
                              \else
                              \refstepcounter{exa\romannumeral\the\@exadepth}
                              \fi
                        \fi
                        \advance\@exadepth \@ne
                       }%
                       {\ifnum \@exadepth =1
                               \fussy\end{list}\normalsize\else {} \fi%
                        \advance\@exadepth \m@ne%
                       }
        \def\@exalabel(#1)(#2)(#3){\makebox[1em][r]{#1}%
                                   \makebox[1em][l]{#2}%
                                   \makebox[1.5em][r]{#3}%
                             }
        \def\@exaitem[#1]{\item[\@exalabel((\arabic{exai})%
            (\ifnum \@exadepth >1%
                    \alph{exaii}%
                    \ifnum \@exadepth >2%
                           \roman{exaiii}%
                    \else{}%
                    \fi%
             \else{}%
             \fi%
            ))(#1)]%
            }
        \def\ex{\refstepcounter{exa\romannumeral\the\@exadepth}%
                \@ifnextchar [{\@exaitem}{\@exaitem[]}%
               }
        \def\exref#1{\expandafter\xdef\csname r@#1\endcsname%
                      {{[\arabic{exai}%
                         \ifnum \@exadepth > 1%
                         \alph{exaii}%
                         \ifnum \@exadepth > 2%
                                \roman{exaiii}%
                         \else{}%
                         \fi%
                         \else{}%
                         \fi%
                         ]}{\thepage}}%
                    }
        \def\exlab#1{\@bsphack\if@filesw {\let\thepage\relax
                     \def\protect{\noexpand\noexpand\noexpand}%
                     \edef\@tempa{\write\@auxout{\string
                     \newlabel{#1}{{(\arabic{exai}%
                                    \ifnum \@exadepth > 1%
                                    \alph{exaii}%
                                    \ifnum \@exadepth > 2%
                                    \roman{exaiii}%
                                    \else{}%
                                    \fi%
                                    \else{}%
                                    \fi%
                                    )}{\thepage}}}}%
                     \expandafter}\@tempa
                     \if@nobreak \ifvmode\nobreak\fi\fi\fi\@esphack}
        \catcode`@=12

\begin{document}


\title{Bridging as Coercive Accommodation}
\author{Johan Bos\thanks{University of the Saarland,
	Postfach 1150, 66041 Saarbr\"ucken, Germany  -- This work was
partly funded by the German Ministry for Research and Technology (BMFT)
under contract 01 IV 101 R.}
\and Paul Buitelaar\thanks{Computer Science, Brandeis University,
	Waltham, MA 02254, USA}
\and Anne-Marie Mineur\thanks{Computerlinguistik, University of
	the Saarland, Postfach 1150, 66041 Saarbr\"ucken, Germany} }
\date{\sf \{bos, mineur\}@coli.uni-sb.de,  paulb@cs.brandeis.edu}

\maketitle


\section{Introduction}
\label{introduction}

\cite{sandt:presupdrt} introduces \cite{lewis}' notion of {\em
accommodation\/} in Discourse Representation Theory \cite{KR:DtoL} as a
tool to account for gaps in the discourse. His theory of presupposition
projection takes presuppositions to behave like anaphora. Anaphoric
expressions normally are linked to antecedents that have previously been
established in the discourse.  If example~\ref{kingoffrance} would
appear in a context where no {\sl king of France} is present -- hence no
antecedent is available -- then Van der Sandt's algorithm {\em
accommodates\/} the existence of a king of France.
	\begin{examples}
	\ex\exlab{kingoffrance}
		When I give a party, the king of France always attends it.
	\end{examples}
This is different from the situation where a definite description can be
{\em linked\/} to an antecedent that was previously introduced by an
indefinite description, as in~\ref{celebrity}. There is no need to
accommodate an antecedent, because there is already a suitable candidate
available.
	\begin{examples}
	\ex\exlab{celebrity}
		When I invite a celebrity, the celebrity never comes.
	\end{examples}
Example~\ref{barkeeper} however is slightly different. There is no
actual antecedent for the anaphoric expression {\sl the barkeeper}, but
because of {\sl a bar}, there isn't really a problem, apparently there
is some implicit antecedent. Van der Sandt's projection algorithm fails
to make this implicit link, and accommodates the existence of {\sl a
barkeeper} to the global context, in fact no theory on presupposition
that we know of can deal with these\footnote{
	The closest comes probably Beaver's dynamic theory of
	presupposition \cite{beaver}. }.
	\begin{examples}
	\ex\exlab{barkeeper}
		When I go to a bar, the barkeeper always throws me out.
	\end{examples}
Contrasting~\ref{barkeeper} with~\ref{playground} makes our point even
clearer; this sentence sounds truly infelicitous. The hearer tries to
somehow link this {\sl barkeeper} with familiar information, and fails.
	\begin{examples}
	\ex[?]\exlab{playground}
		When I go to a playground, the barkeeper always throws me out.
	\end{examples}
{\sl A bar} provides sufficient information to license {\sl the
barkeeper}, but in {\sl a playground} there is nothing that can establish
such a link. Making a link between the new discourse referent (i.e. {\sl
the barkeeper}) to the network of discourse referents that is already
established, is called {\em bridging\/} (\cite{bridging},
\cite{heim:diss}). Definite descriptions that can be bridged to existing
information do not need the accommodation of new referents;
example~\ref{kingoffrance} requires accommodation, but \ref{barkeeper}
can be solved with mere bridging. An adequate theory of presupposition
obviously needs a serious explanation of bridging to account for the
projection problem of presupposition.\\

To account for these phenomena, we borrow from \cite{pus:GL} and compare
{\em bridging\/} with {\em coercion\/}. Pustejovsky presents examples
like~\ref{beginabook}:
	\begin{examples}
	\ex\exlab{beginabook}
		I would like to begin a new book tonight.
	\end{examples}
Here, too, some information is missing: {\sl begin} implies some event,
but {\sl a new book} is an artifact. The fact that the speaker should be
interpreted as {\sl beginning to read the book}, or -- if he is a writer
-- {\sl to write} one, is motivated by what we know about {\sl book}.
Pustejovsky claims that such information should be considered lexical
knowledge of the noun, which is represented in a so-called {\em qualia
structure\/}. Based on this information, arguments of improper types can
be {\em coerced\/} to proper ones. We will see how a similar approach
can be followed to account for {\em bridging\/}.

In section~\ref{qscoercion} we will present Pustejovsky's ideas in more
detail, explaining concepts like {\em coercion\/} and {\em qualia
structure\/}. In Pustejovsky's work these ideas only get applied on the
sentence level. Section~\ref{bridging} will show how the ideas of Van
der Sandt and Pustejovsky fit together very nicely, even complementing
each other and we will show that {\em bridging\/} operates intra- as well
as inter-sentential. In section~\ref{examples} we will present some
examples of {\em linking\/}, {\em bridging\/} and {\em accommodation\/},
and in section~\ref{functionalcomposition} we will discuss the notion of
functional composition and coercion in this model.


\section{Qualia Structure and Coercion}
\label{qscoercion}

\subsection{Qualia Structure}
\label{qscoercionqs}

In \cite{pus:GL} and subsequent papers the notions of {\em coercion} and
{\em qualia structure} have been introduced. Qualia structure can be seen
as a set of lexical entailments. For instance, the word {\sl book}
entails at least the two events of {\sl reading} and {\sl writing} it,
besides the knowledge that it consists of several separate parts, like
the {\sl cover}, {\sl pages}, etc. Pustejovsky suggests four {\em qualia
roles} to represent such knowledge: {\sc formal}, {\sc constitutive},
{\sc telic} and {\sc agentive}. In \cite{pus:GL} these have been
defined as follows\footnote{
	In more common AI-related terms we could rephrase them as:
	{\sc formal} - {\sc isa}, {\sc constitutive} - {\sc part-of~/~hasa}, {\sc
telic} - {\sc purpose} and {\sc agentive} - {\sc
	cause}.}:

\begin{itemize}
\item {\sc formal}: That which distinguishes the object within a larger
domain.
\item {\sc constitutive}: The relation between an object and its
constituents or proper parts.
\item {\sc telic}: Purpose and function of the object.
\item {\sc agentive}: Factors involved in the origin or "bringing about"
of an object.
\end{itemize}

\label{important}
The exact structure of this kind of lexical semantic knowledge seems to be
very intricate. Again for the same example, it is important for instance
to realize that a {\sl book} is at the same time a {\em physical object}
and an {\em information container}. The first description considers the
physical viewpoint, whereas the second defines the conceptual angle of
what constitutes our idea of a {\sl book}. What angle one takes ({\em
physical object} or {\em information container}) has immediate consequences
for the knowledge that is represented in the rest of the qualia structure.
The composing parts of the physical side of a book ({\sl pages}, {\sl cover},
etc.) are different from those of the conceptual side ({\sl title}, {\sl
sections}, {\sl paragraphs}, etc.). The same goes for the representation
of typical events a book is involved in. The physical 'quality' (qual) of
a book can be {\sl printed}, {\sl typeset} or even {\sl shelved}. The
information 'quality' can be said to undergo the events of {\sl reading},
{\sl writing} as mentioned before. It is however undeniable that the
two main qualities of book, along with all their entailments, are intimately
related to each other and should be represented accordingly in one
comprehensive (qualia) structure.

\subsection{Coercion}
The need for a rich lexical semantic knowledge representation like qualia
structure becomes clear in considering sentences like \ref{beginabook}
above, which is repeated here:
	\begin{examples}
	\ex\exlab{beginabook2}
		I would like to begin a new book tonight.
	\end{examples}
As mentioned before, the verb {\sl begin} expects an event here but has to
settle for an {\sl artifact} (book). We can now use the qualia structure of
this {\sl artifact} to infer some {\sl event} that is entailed by it and
which can stand in its place. This is an example of what Pustejovsky has
called {\em metonymic reconstruction} \cite{pus:GL} for cases where an
interpretation can be inferred from some partial meaning of the word in
question. In more general terms, anytime a word or phrase is not of the
desired type\footnote{
	Possibly this use of the term {\em type} is not appropriate and
	we should use {\em sort} instead. However, here we present the
	terms as they have been defined by \cite{pus:GL}.}
(like {\sl artifact}, {\sl event}, etc.) we are allowed to {\em coerce} it
into one of its entailments that is of the appropriate type, where the
entailments are stored in its qualia structure. Another example of this is
the following sentence:
	\begin{examples}
	\ex\exlab{announcenewmodel}
		BMW announced a new model.
	\end{examples}
Here the verb {\sl announce} is looking for a subject of type {\em animate}
while only one of type {\em institute} is available. The qualia structure of
any {\em institute} however should represent the fact that they are made up
of people, which are {\em animate} entities. So, in this sentence we can
infer that some human at the BMW company did the actual announcement.\\

This summarizes Pustejovsky's program as described in \cite{pus:GL} and
subsequent papers. In this paper we extend coercion with the notion of
context, which seems not only a valid research topic but also
desperately needed because of the restricted explaining power of
coercion if context is not considered. Take for instance
sentence~\ref{beganabook} :
	\begin{examples}
	\ex\exlab{beganabook}
		John began a book.
	\end{examples}
Although above we assumed several times that one can infer {\sl read} and
{\sl write} events from the qualia structure of {\sl book} in order to make
this sentence semantically well formed, this can only be a default
approximation. We would need an actual context for this sentence to decide
what event exactly should be inferred. Imagine for instance a dinner,
organized by the {\sc literary and culinary society}, where all dishes are
shaped in the form of books$\ldots$\footnote{
        Still another problem concerning the lack of context is
        illustrated by the following examples where no argument
        at all is available for coercion to take place, {\sl Monday} and
	{\sl yesterday} are modifiers:
        \begin{quote}
        I propose Monday.\\
	I began yesterday.
        \end{quote}
        Pustejovsky (personal communication) has termed this
        loosely as {\em null coercion}, because although coercion
        should take place it cannot be executed properly. Taking
        context into account could be of help however to make the
	sentences sound more natural, as the following examples show:
        \begin{quote}
        Let's make an appointment. I propose Monday.\\
	Let's play darts. I begin.
        \end{quote}
	It seems that null coercion should coerce an anaphor
	which is of the required type. In both examples this would
	be event-type anaphors. }

This example is farfetched, but it may make the point more clear. We do
not assume that the qualia structure of {\sl book} should contain any
reference to this particular example. It is important however to realize
that any {\sl artifact} entails by default a number of events in which
it is engaged. In this particular context these events would be
overruled.


\section{Bridging in DRT with Qualia Structure}
\label{bridging}

This section shows how we deal with anaphora resolution in general,
and particularly bridging, in a version of DRT which uses extensively
{\em qualia\/} information. We define the language of
Discourse Representation Structures (DRSs) of our extended DRT,
show how resolution works, and finally give some detailed examples.

\subsection{A Sketch of the Architecture}

Basically, we extend Van der Sandt's theory of presupposition
with the notion of bridging anaphora. In short, Van der Sandt
views presupposition as anaphora with more descriptive content,
and uses one and the same mechanism for dealing with
both phenomena \cite{sandt:presupdrt}. Anaphoric information
can either be resolved to an antecedent that is available
from discourse, or if no antecedent is found, be accommodated.
We add a possibility of bridging to the resolution algorithm.
The basic architecture of the system is:

\begin{enumerate}
\item parse sentence: result is a sentence-DRS
\item merge sentence-DRS with main-DRS
\item perform anaphora resolution
\end{enumerate}

A {\em sentence-DRS} is a DRS with all anaphoric information
unresolved, and is the result of a bottom-up driven semantic
construction dependent on some syntactic structure.
A sentence-DRS can be viewed as a sort of under-specified
logical form with respect to anaphoric information.
Special types of DRSs ($\alpha$-DRSs) mark anaphoric
information. The {\em main-DRS} is the DRS of the context
interpreted so far. It is a {\em proper} DRS, i.e., a DRS with
no unresolved anaphoric information. Proper DRSs can be
interpreted as in standard DRT: they are {\em true} with respect to
a certain model if they can be {\em embedded\/} in that model
(\cite{KR:DtoL}. Before we explain how anaphora resolution works we
define DRSs and the merging operation.

\subsection{Discourse Representation Structures}

Let's introduce  some terminology.
Discourse markers are variables ranging over objects in
the domain. Terms are either discourse markers or
DRSs. Furthermore,
we adopt a typed lambda-calculus for DRSs \cite{tilburg,muskens:cdrt}.
DRSs are defined as follows:

\begin{quote}
{\bf Definition 1. DRS}\\
If $U$ is a set of discourse markers,
$C$ is a set of DRS-Conditions, and t$_1$,...,t$_n$ terms, then
$<U,C>$ is a DRS,
$<U,C> \oplus <U^\prime,C^\prime>$ is a DRS,
$\lambda$ t$_1$,...,t$_n$. $<U,C>$ is a DRS. Nothing else is a DRS.
\end{quote}

\begin{quote}
{\bf Definition 2. DRS-Conditions}\\
If x$_1$,..,x$_n$ are discourse markers, P an n-place
condition, K and K$_1$ DRSs, then
P(x$_1$,...,x$_n$),
x$_1$ = x$_2$,
K $\to$ K$_1$,
$\lnot$ K,
K $\lor$  K$_1$,
$\alpha$:K, and
Q:K are DRS-Conditions. Nothing else is a DRS-Condition.
\end{quote}

The first five DRS-Conditions we already know from standard DRT
\cite{KR:DtoL} and need no further explanation. So called $\alpha$-DRSs
represent unresolved anaphoric information. DRSs that contain
$\alpha$-DRSs are therefore unresolved DRSs.  Q-DRSs represent qualia
structure, with Q$_F$ for {\sl formal}, Q$_C$ for constitutive, Q$_A$
for agentive and Q$_T$ for telic. For notational purposes we use
$\cal{Q}$ to represents a {\em set\/} of qualia-DRSs\footnote{
	As mentioned in section~\ref{qscoercionqs} on
	page~\pageref{important}, the distinction that is made in the
	formal role carries through in all other qualia roles. This
	could be represented by embedding Q$_C$, Q$_A$ and Q$_T$ in
	Q$_F$. This is beyond the scope of our paper.}. Now for merging:

\begin{quote}
{\bf Definition 3. Merging ($\oplus$)}.\\
$<U_1,C_1> \oplus <U_2,C_2> = <U_1 \cup U_2, C_1 \cup C_2>$
\end{quote}

The merge operation takes two DRSs and makes a union of the
sets of discourse markers and a union of the sets conditions.
Merging of DRSs is used both for constructing DRSs
(cf. \cite{tilburg}) and {\em coercive accommodation}.

The latter term brings us to the next definition.
Qualia-information, represented
in Q-DRSs is normally not
accessible and does not affect the truth-conditions of a DRS.
It is introduced in the lexicon and brought into discourse via
the DRS bottom-up construction algorithm. If necessary, for example
to play the role of antecedent, the
qualia structure is put forward to the surface by a process we
call {\em coercive accommodation}. It is defined as a function from
DRSs to sets of DRSs:

\begin{quote}
{\bf Definition 4. Coercive Accommodation (CA).}\\
{\sc ca}($<U,C>$) = $\{ <U,C> \oplus$ K $|$ Q:K $\in C\}$
\end{quote}

Note that CA is always local: it cannot accommodate qualia
information which is embedded. Note also that we have defined
CA only for DRS without lambda's: this will do for the
purposes of this paper.
Q-DRSs are also used for {\em type coercion}, which is
discussed later on in this paper.

In DRT the structure of DRSs restricts the choice of possible
antecedents of an anaphoric construction. For a discourse marker
to be the antecedent for an anaphor, it must be {\em accessible}
from the DRS which the anaphor is represented.
To define accessibility of DRSs and discourse markers
we first use the notion {\em  subordination} between DRSs.
We adopt the notation C(K) meaning the set of conditions of
DRS (K), and U(K) meaning  the set of discourse markers of K.

\begin{quote}
{\bf Definition 5. Subordination.}\\
If K$_1$, K$_2$, and K$_3$ are DRSs, then K$_2$ is subordinated
to K$_1$ (or K$_1$ subordinates K$_2$) if
K$_1$ $\oplus$ K$_2$,
K$_1$ $\to$ K$_2$ $\in$ C(K$_3$),
K$_2$ $\to$ K$_3$ $\in$ C(K$_1$),
$\lnot$ K$_2$ $\in$ C(K$_1$),
K$_2$ $\lor$ K$_3$ $\in$ C(K$_1$),
K$_3$ $\lor$ K$_2$ $\in$ C(K$_1$),
$\alpha$:K$_2$ $\in$ C(K$_1$),
Q:K$_2$ $\in$ C(K$_1$), and
K$_2$ is subordinated to K$_3$ and K$_3$ is subordinated to K$_1$.
\end{quote}

So, if x is a discourse marker and K$_1$ and K$_2$ are DRSs,
and x is in the domain of K$_1$ (x $\in$ U(K$_1$)),
then x is accessible from K$_2$ if K$_2$ is subordinated to K$_1$.

\subsection{Anaphora Resolution}

Left to explain is how anaphora resolution works.
We repeat for convenience that resolution can take place in three
different ways:
\begin{enumerate}
\item resolution to an accessible, suitable discourse marker ({\em linking})
\item resolution to coercively accommodated material of an
      accessible DRS ({\em bridging})
\item accommodation of the anaphoric information to an
     accessible DRS ({\em accommodation})
\end{enumerate}

We introduced accessibility already,
but haven't explained yet
the notion of `suitable' discourse marker, or
better: suitable DRSs. Suitability is an extra constraint
on the choice of antecedent. A DRS is suitable to another
DRS if there is a way you find a match between discourse
markers and conditions between both. More formally:

\begin{quote}
{\bf Definition 6. Suitability.}\\
A DRS K$_2$ is $m$-suitable to DRS K$_1$ if there is a mapping
$m$ such that scope($m$)=U(K$_2$) and for every x it is the
case that m(x) $\in$ U(K$_1$) and there is a DRS K$_3$ such that
C(K$_3$) $\subseteq$ C(K$_1$) if U(K$_3$) = $\{$ m(x) $|$ x $\in$
U(K$_1$)$\}$.
\end{quote}

We now introduce the heart of the system: anaphora resolution.
This algorithm works as follows. All anaphoric information in
the main-DRS (of course after merging it with the sentence-DRS of
the last processed sentence) is resolved. This information
is clearly marked because these are just our $\alpha$-DRSs. Resolution
either unifies this material with a suitable antecedent or
accommodates it, and as a result, $\alpha$-DRSs disappear.
After resolving all $\alpha$-DRSs, we are left with a proper-DRS,
a DRS which is fully specified with respect to anaphoric information.
This DRS is model-theoretically interpretable, as in standard DRT.

To describe the component,
we use K$_\alpha$ to indicate anaphoric DRSs, and K$_m$ for
the main-DRS. Definition 7 describes  a function that takes a
certain main-DRS and a certain $\alpha$-DRS from it, and
returns a set of DRSs (since there could be more
than one possible antecedent or accommodation site)
with this $\alpha$-DRS resolved. The output of this function
can be fed back into the same function until all anaphoric
information is resolved (all $\alpha$-DRS have been
consumed).\footnote{The order of which resolution of anaphoric structure
takes place is important as well. We don't pay any attention to this,
but see \cite{sandt:presupdrt}.}

\begin{quote}
{\bf Definition 7. Anaphora Resolution.}\\
\begin{tabular}{ll}
{\sc ar}(K$_\alpha$,K$_m$)
& =  $\{$ K' $|$  K' $\in$  {\sc link}(K$_\alpha$,K$_m$) $\}$ iff
     $|${\sc link}(K$_\alpha$,K$_m$)$|$ $>$ 0\\
& =  $\{$ K' $|$  K' $\in$  {\sc bridge}(K$_\alpha$,K$_m$) $\}$ iff
     \parbox[t]{44mm}{$|${\sc link}(K$_\alpha$,K$_m$)$|$ $=$ 0 and\\
                  $|${\sc bridge}(K$_\alpha$,K$_m$)$|$ $>$ 0}\\
& =  $\{$ K' $|$  K' $\in$  {\sc acc}(K$_\alpha$,K$_m$) $\}$ iff
     \parbox[t]{44mm}{$|${\sc link}(K$_\alpha$,K$_m$)$|$ $=$ 0 and\\
                      $|${\sc bridge}(K$_\alpha$,K$_m$)$|$ $=$ 0}\\
\end{tabular}
\end{quote}

Note that this definition prefers linking to bridging, and
bridging to accommodation, which we assume is right.
{\sc link}, {\sc bridge}, and {\sc acc} are functions from
the main DRS to sets of DRSs. We use DRSsubstitution to
describe these operations ($[$ K$_1$ / K$_2$ $]$ K$_3$ means
that K$_1$ is substituted for K$_2$ in K$_3$).

\begin{quote}
{\bf Definition 8. Linking.}\\
{\sc link}(K$_\alpha$,K$_m$) = $\{$ \parbox[t]{100mm}{
  $[$ K$_3$ / K$_2$ $]$ K$_m$  $|$
   K$_\alpha$ is subordinated and m-suitable to K$_1$ \&
   $\alpha$:K$_\alpha$ $\in$  C(K$_2$) \&
   U(K$_3$)=U(K$_2$) $\cup$ U(K$_\alpha$) \&
   C(K$_3$)=C(K$_2$)-$\alpha$:K$\alpha$ $\cup$ C(K$_\alpha$) $\cup$
     $\{$ x=y $|$ m(x)=y $\}$
   $\}$ }
\end{quote}

\begin{quote}
{\bf Definition 9. Bridging.}\\
{\sc bridge}(K$_\alpha$,K$_m$) = $\{$ \parbox[t]{100mm}{
   K'$_m$~$|$~K$_\alpha$ is subordinated K$_4$ \&
   K$_1$ $\in$ {\sc ca}(K$_4$) \& \linebreak
   m-suitable to K$_1$ \& $\alpha$:K$_\alpha$ $\in$  C(K$_2$) \& \linebreak
   U(K$_3$)=U(K$_2$) $\cup$ U(K$_\alpha$) \&
   C(K$_3$)=C(K$_2$)-$\alpha$:K$\alpha$ $\cup$ C(K$_\alpha$) $\cup$
     $\{$~x=y~$|$~m(x)=y~$\}$ \&
   K'$_m$ = $[$ K$_3$ / K$_2$ $]$ K$_m$ \&
   K'$_m$ = $[$~K$_1$~/~K$_4$~$]$~K$_m$
   $\}$ }
\end{quote}

\begin{quote}
{\bf Definition 10. Accommodation.}\\
{\sc acc}(K$_\alpha$,K$_m$) = $\{$ \parbox[t]{100mm}{
     K'$_m$ $|$
    K$_\alpha$ is subordinated to K$_1$ \&
    $\alpha$:K$_\alpha$ $\in$ C(K$_2$) \&
    U(K$_3$)=U(K$_2$) \&
    C(K$_3$)=C(K$_2$)-$\alpha$:K$_\alpha$ \&
    K'$_m$=$[$~K$_1$~$\oplus$~K$_\alpha$~/~K$_1$ $]$ K$_m$ \&
    K'$_m$=$[$~K$_3$ / K$_2$ $]$ K$_m$~$\}$ }
\end{quote}

Accommodation has its limits. First, it shouldn't introduce
free variables, and Van der Sandt introduces a number
of acceptability rules for accommodation. These are
briefly: resolution should not introduce contradictions and
require a contribution to discourse. For more discussion
on this issue the interested reader  should consult
\cite{sandt:presupdrt}.


\subsection{Examples of Linking, Bridging and Accommodation}
\label{examples}

This section exemplifies the notions {\em linking}, {\em bridging}, and
{\em accommodation}, which we introduced in the previous section.
We will do this in view of the examples given in the introduction.
For each of these examples we give the DRS with all
anaphoric information unresolved, and the fully resolved derived
after anaphora resolution as well. For reasons of clarity, only
the relevant parts of the DRSs are deeply analyzed.

\subsubsection{Linking}

The first example involves simple linking between anaphor and
antecedent. Consider the unresolved DRS of \ref{celebrity}:

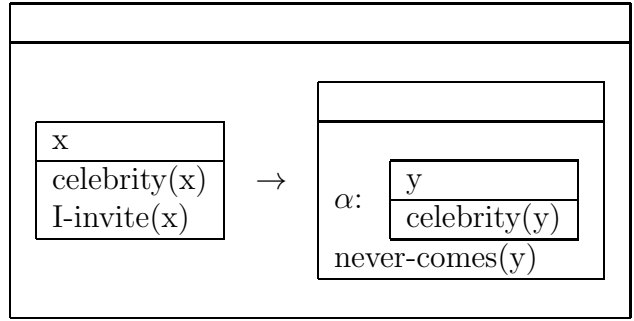
\begin{center}
\drs{}{\\
   \drs{x}{celebrity(x)\\
          I-invite(x)} $\to$
   \drs{}{\\$\alpha$:\drs{y}{celebrity(y)}\\
          never-comes(y)}\\}
\end{center}

The definite description introduces an $\alpha$-DRS for {\sl the
celebrity}, since this is presupposed information.
Trying to link this anaphoric information is successful,
since there is an accessible suitable discourse marker available.
The result is the resolved DRS:

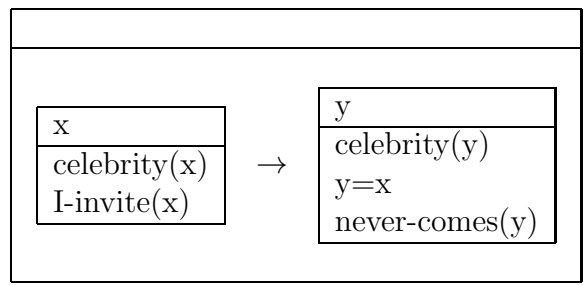
\begin{center}
\drs{}{\\
   \drs{x}{celebrity(x)\\
          I-invite(x)} $\to$
   \drs{y}{celebrity(y)\\
          y=x\\
          never-comes(y)}\\}
\end{center}

This DRS can be read as: {\sl If I  invite a celebrity, he never comes}.

\subsubsection{Bridging}

Now for our bridging\footnote{
        Bridging does not seem to be the preferred option in the case
        of resolution of pronouns as the following examples show:

        \begin{quote}
        When I go to a bar, he always throws me out.\\
        When BMW announced a new model, he looked very proud.
        \end{quote}

        In both sentences a reading for 'he' can be found by linking
        to coercively accommodated material out of the Q-DRS from
        respectively {\sl bar} and {\sl BMW}, i.e. {\sl a barkeeper} or
	{\sl a spokesperson}. However they don't seem to be the
	preferred readings as has been shown by
	\cite{McGlashan,SanfordandGarrod} for similar examples.}
	example.
The unresolved DRS of example~\ref{barkeeper} is (simplifying the Q-DRS
for convenience):

\begin{center}
\drs{}{\\
   \drs{x}{bar(x)\\
          Q:\drs{z}{barkeeper(z)\\of(z,x)}\\
          I-go-to(x)} $\to$
   \drs{}{\\
      $\alpha$:\drs{y}{barkeeper(y)}\\
          always-throws-me-out(y)}\\}
\end{center}

The presupposition trigger {\sl the barkeeper} introduces the
anaphoric information. Linking fails, the only available
discourse marker is not suitable since the condition of anaphoric
information does not match with it. Bridging is successful, though,
yielding the resolved DRS:

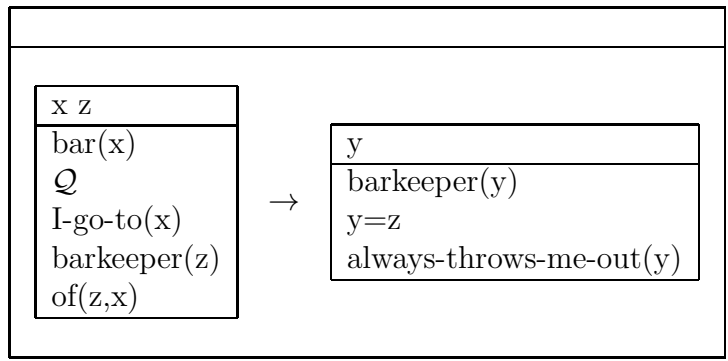
\begin{center}
\drs{}{\\
   \drs{x z}{bar(x)\\
          $\cal{Q}$\\
          I-go-to(x)\\
          barkeeper(z)\\
          of(z,x)} $\to$
   \drs{y}{barkeeper(y)\\
      y=z\\
      always-throws-me-out(y)}\\}
\end{center}

This DRS does not assume a particular barkeeper that throws
the speaker out, but a barkeeper that belongs to the bar the
speaker goes to -- the correct prediction.

\subsubsection{Accommodation}

Accommodation is our emergency case: {\em if everything fails, then
accommodate}. This happens in cases like \ref{kingoffrance}, which
unresolved DRS is:

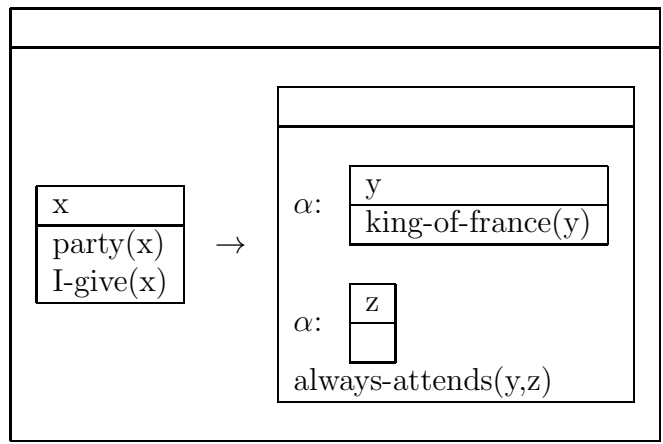
\begin{center}
\drs{}{\\
   \drs{x}{party(x)\\
          I-give(x)} $\to$
   \drs{}{\\
        $\alpha$:\drs{y}{king-of-france(y)}\\ \\
          $\alpha$:\drs{z}{}\\
          always-attends(y,z)}\\}
\end{center}

The pronoun represented by the discourse marker {\sl z} can be linked to
{\sl x}. But we cannot link {\sl the king of France} to some accessible
discourse marker, nor is there a way to make bridging inference. The
only possibility left is to accommodate the king:

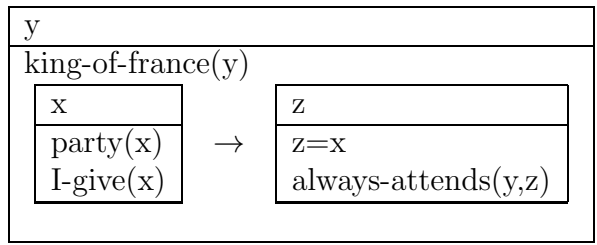
\begin{center}
\drs{y}{king-of-france(y)\\
   \drs{x}{party(x)\\
          I-give(x)} $\to$
   \drs{z}{z=x\\
      always-attends(y,z)}\\}
\end{center}

This DRS represents the reading: {\sl there is a king of France, and if
I give a party, he will attend it}. This is again the correct
prediction.


\section{Functional Composition and Coercion}
\label{functionalcomposition}

\subsection{Defining the notions}
Functional Composition, including type coercion,
is defined as follows, K$_1$ being the functor,
K$_2$ the argument, and $\vec{\sigma}$ a sequence of
terms such that K$_2$(t)($\vec{\sigma}$) is a proposition:

\begin{quote}
{\bf Definition 11. Functional Composition ($\odot$).}\\
\begin{tabular}{ll}
K$_1$ $\odot$ K$_2$ &
=    $\lambda$ $\vec{\sigma}$.
    K$_1$($\lambda$ v. (K$_2$(v)($\vec\sigma$)))
          iff K$_1$ is of type $<\alpha$,t$>$ and v is of type $\alpha$; \\
 & = K$_1$ $\odot$ K$_3$ (where K$_3$ $\in$ {\sc tc}(K$_2$)) otherwise.
\end{tabular}
\end{quote}

Clause one is like the functional composition rule (in
\cite{tilburg}). This rule has the nice property that
it doesn't need type-shifting of arguments. It always binds
the first argument position of the argument, and has functional
application as a special case ($\vec{\sigma}$ is empty then).
The second clause does the type coercion stuff (cf. \cite{pus:TC}):

\begin{quote}
{\bf Definition 12. Type Coercion (TC).}\\
{\sc tc}(K)=$\{$ K$^\prime$ $\odot$ K $|$ K$^\prime$ $\in$ {\sc qa}(K) $\}$
\end{quote}

\begin{quote}
{\bf Definition 13. Qualia Access (QA)}.\\
{\sc qa}(K)=$\{$ K$_Q$ $|$ Q:K$_Q$ $\in$ C(K)
         or {\sc qa}(K') where K' is a sub-DRS of K $\}$
\end{quote}

Note that {\sc tc} also works for arbitrarily deep embedded DRSs by use
of the Qualia Access function. This is nice for quantified NPs like {\sl
every book}, where the qualia DRSs lexically introduced for {\sl book}
has been placed in the restrictor.

\subsection{Some lexical entries}

In this section we present some example lexical entries. In this paper we
will only assign (a simplified) qualia structure to nouns\footnote{
	\cite{pus:book} assumes qualia structures for {\em all}
	categories.},
see {\bf book}. Lexical entries can be abbreviated by their boldface
notation -- {\bf write} stands for the semantic part of the
lexical entry of {\sl write}. We use small e,x,y and z for variables
over type e (for entities, i.e. objects and events), capital P for
DRS of type $<$e,t$>$ (properties), and capital E for event-types
(normally $<e,t>$).

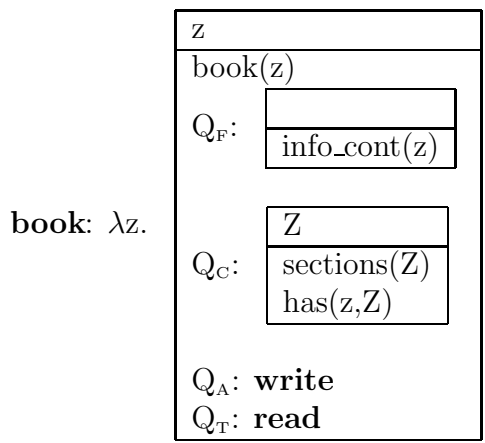
\begin{tabbing}
{\bf pres} \= : \= \kill
{\bf book}\>:\> $\lambda$z.\drs{z}{book(z)\\
         		    \qdrs{F}{}{info\_cont(z)}\\ \\
			    \qdrs{C}{Z}
                                     {sections(Z)\\
				      has(z,Z)}\\ \\
			     Q$_{\mbox{\tiny A}}$: {\bf write} \\
			     Q$_{\mbox{\tiny T}}$: {\bf read}}
\end{tabbing}

By introducing determiners ({\bf a, the, every}) we account for the
possibility to carry qualia structure through the derivation. Note the
difference between these three determiners. The article {\em the}
introduces an $\alpha$-DRS since it is a presupposition trigger.

\begin{tabbing}
{\bf pres} \= : \= \kill
{\bf a} \>:\>
  $\lambda$ P$_1$ P$_2$. \drs{x}{} $\oplus$ P$_1$(x) $\oplus$ P$_2$(x)\\ \\
{\bf the} \>:\>
  $\lambda$ P$_1$ P$_2$.
  \drs{}{\mbox{}\\[-2ex] $\alpha$: \drs{x}{} $\oplus$ P$_1$(x)\mbox{}\\[-2ex]}
$\oplus$ P$_2$(x)\\ \\
{\bf every} \>:\>
  $\lambda$ P$_1$ P$_2$.
  \drs{}{\mbox{}\\[-2ex]\drs{x}{} $\oplus$ P$_1$(x) $\to$
P$_2$(x)\mbox{}\\[-2ex]}
\end{tabbing}

The proper name {\bf john} introduces an anaphoric DRS
which is merged with the representation of its predicate. Proper names
do {\em not\/} have qualia structure (see footnote earlier).

\begin{tabbing}
{\bf pres} \= : \= \kill
{\bf john}\>:\>
$\lambda$ P. \drs{}{\mbox{}\\[-2ex]$\alpha$:\drs{x}{john(x)}\mbox{}\\[-2ex]}
$\oplus$ P(x)
\end{tabbing}

The verbs 
{\bf write} and {\bf read} introduce event-types. Lambda-operators bind the
variables that will fulfill the thematic roles {\em agent\/} and {\em theme}.

\begin{tabbing}
{\bf pres} \= : \= \kill
{\bf write} \>:\> $\lambda$ y x e. \drs{}{write(e)\\
				          agent(e,x)\\
                                          theme(e,y)}\\ \\
{\bf read} \>:\> $\lambda$ y x e. \drs{}{read(e)\\
                                         agent(e,x)\\
                                         theme(e,y)}
\end{tabbing}

The aspectual verb
{\bf begin} expects something that expresses an event-type.
We here simply treat it as a modifier, and ignore its further
aspectual presuppositions.

Finally,  tense {\bf pres}
applies to an event-type and binds off the event variable:the
result is a DRS of type t, i.e. a DRS with no lambda variables.

\begin{tabbing}
{\bf pres} \= : \= \kill
{\bf begin} \>: \> $\lambda$ E x e. \drs{}{begin(e)} $\oplus$ E(x)(e) \\ \\
{\bf pres} \> : \> $\lambda$ E. \drs{e}{now(e)} $\oplus$ E(e)
\end{tabbing}

\subsection{A sample derivation}

Let us now follow the derivation of `John begins a book'. Functional
composition of {\bf a} with {\bf book} yields a noun phrase that
contains the qualia structure of the noun and awaits a property to
merge with.

\begin{quote}
{\bf a} $\odot$ {\bf book}   =
$\lambda$P.\drs{z}{book(z)\\
         		    \qdrs{F}{}{info\_cont(z)}\\ \\
			    \qdrs{C}{Z}
                                     {sections(Z)\\
				      has(z,Z)}\\ \\
			     Q$_{\mbox{\tiny A}}$: {\bf write} \\
			     Q$_{\mbox{\tiny T}}$: {\bf read}}
$\oplus$ P(z)
\end{quote}

Functional composition of {\bf begin} with {\bf a $\odot$ book} can only
work with a type coercion. The event that {\bf begin} requires cannot be
found directly, so no simple link can be made. From the qualia structure
of {\bf a $\odot$ book} we can for example coerce {\bf read}, and
this qualifies as the required event.
This coercion step is worked out later.

\begin{quote}
{\bf begin} $\odot$ ({\bf a} $\odot$ {\bf book})  =
   $\lambda$ x. e.
   \drs{y}
       {begin(e)\\
        read(e)\\
        agent(e,x)\\
        theme(e,y)\\
        book(y)\\
        $\cal{Q}$}
\end{quote}

The rest of the derivation follows straightforwardly. {\bf begin $\odot$
(a $\odot$ book)} functionally composed with {\bf john} results in a
lambda-DRS.

\begin{quote}
{\bf john} $\odot$ ({\bf begin} $\odot$ ({\bf a} $\odot$ {\bf book}))  =
   $\lambda$ e.
  \drs{y}
	{\mbox{}\\[-2ex]$\alpha$:\drs{x}{john(x)\mbox{}\\[-2ex]}\\
	begin(e)\\
	read(e)\\
	agent(e,x)\\
	theme(e,y)\\
	book(y)\\
        $\cal{Q}$
}
\end{quote}

Adding tense ({\bf pres}) to the lambda-DRS turns it into a proper
DRS -- all anaphoric information has been resolved and no antecedent for
the presupposed event needed to be accommodated.

\begin{quote}
{\bf pres} $\odot$ ({\bf john} $\odot$ ({\bf begin} $\odot$ ({\bf a} $\odot$
{\bf book})))  =
  \drs{e y}{\mbox{}\\[-2ex]$\alpha$:\drs{x}{john(x)\mbox{}\\[-2ex]}\\
	now(e)\\
        begin(e)\\
	read(e)\\
	agent(e,x)\\
	theme(e,y)\\
	book(y)\\
        $\cal{Q}$}
\end{quote}

\pagebreak
Naturally we could just as easily have taken {\bf write} instead of {\bf
read}, or for that matter, any of the other events that occur in the
qualia structure of {\bf book}. Since {\bf read} and {\bf write} are the
only events, the result of coercing {\bf a book} is as follows:\\

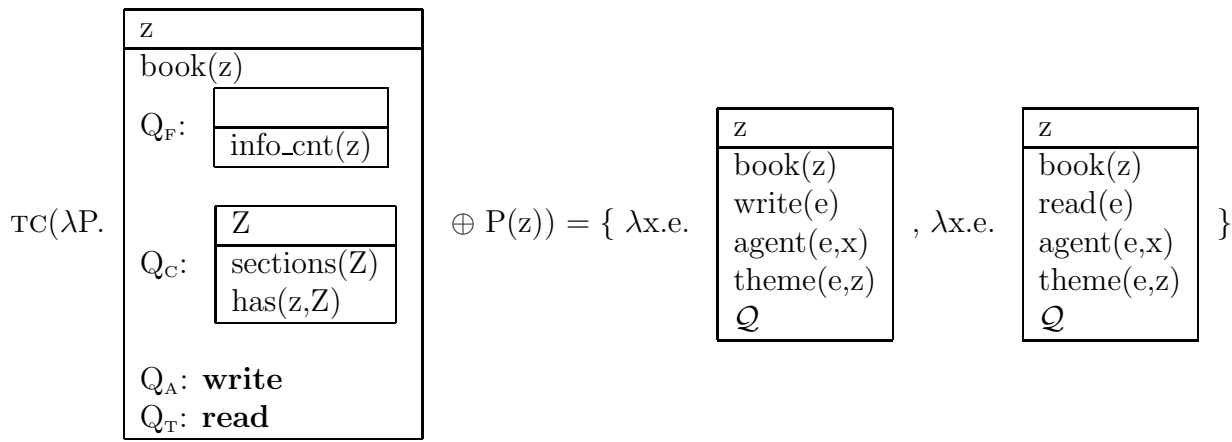
\begin{minipage}{\textwidth}
{\sc tc}($\lambda$P.\drs{z}{book(z)\\
         		    \qdrs{F}{}{info\_cnt(z)}\\ \\
			    \qdrs{C}{Z}
                                     {sections(Z)\\
				      has(z,Z)}\\ \\
			     Q$_{\mbox{\tiny A}}$: {\bf write} \\
			     Q$_{\mbox{\tiny T}}$: {\bf read}}
$\oplus$ P(z)) =
$\{$ $\lambda$x.e.\drs{z}
                       {book(z)\\
			write(e)\\
			agent(e,x)\\
			theme(e,z)\\
			$\cal{Q}$},
     $\lambda$x.e.\drs{z}{book(z)\\
			read(e)\\
			agent(e,x)\\
			theme(e,z)\\
			$\cal{Q}$}$\}$
\end{minipage}


\section{Conclusions and Further Work}
\label{conclusions}

We have shown that Bridging and Coercion can be seen in very much the
same light, viz. as using implicit lexical information to accommodate
a missing antecedent. In doing so, we have extended Pustejovsky's ideas
on Coercion and placed it in a discourse perspective. On the other hand
we have extended Van der Sandt's algorithm with Bridging, and thus made
it more complete with respect to the linguistic data.

The work presented here is limited to definite descriptions; we have not
looked into other presupposition triggers. \cite{beaver} mentions the
following examples, where inferencing takes place.
	\begin{examples}
	\ex\exlab{beaver1}
		Probably, if Jane takes a bath, Bill will be annoyed
		that there is no more hot water.
	\ex\exlab{beaver2}
		If Spaceman Spiff lands on Planet X, he will be bothered
		by the fact that his weight is higher than it would be
		on Earth.

	\end{examples}
In \ref{beaver1} the inference is made that taking a bath uses up a hot
water reservoir, in \ref{beaver2} that landing on a strange planet may
make changes to your weight. To fit with these examples in the framework
we presented in this paper remains for future research.



\begin{thebibliography}{}

\bibitem[\protect\citeauthoryear{Beaver}{1993}]{beaver}
David Beaver.
\newblock What Comes First in Dynamic Semantics.
\newblock manuscript, November 1993.

\bibitem[\protect\citeauthoryear{Bos \bgroup \em et al.\egroup
  }{1994}]{tilburg}
J.~Bos, E.~Mastenbroek, S.~McGlashan, S~Millies and M.~Pinkal.
\newblock A Compositional DRS-based Formalism for NLP-Applications.
\newblock In Harry Bunt, Reinhard Muskens and Gerrit Rentier (eds.), {\em
  International Workshop on Computational Semantics}. Tilburg University, The
  Netherlands, December 1994.

\bibitem[\protect\citeauthoryear{Clark}{1975}]{bridging}
H.H. Clark.
\newblock Bridging.
\newblock In R~Schank and B~Nash-Webber (eds.), {\em Theoretical Issues in
  National Language Processing}. MIT, June 1975.

\bibitem[\protect\citeauthoryear{Heim}{1982}]{heim:diss}
Irene Heim.
\newblock {\em The Semantics of Definite and Indefinite Noun Phrases}.
\newblock PhD thesis, University of Massachusetts at Amherst, 1982.
\newblock (Sonderforschungsbereich 99 Linguistik, Universit{\"a}t Konstanz).

\bibitem[\protect\citeauthoryear{Kamp and Reyle}{1993}]{KR:DtoL}
Hans Kamp and Uwe Reyle.
\newblock {\em From Discourse to Logic; introduction to modeltheoretical
  semantics of natural language, formal logic, and discourse representation
  theory}.
\newblock Kluwer, 1993.

\bibitem[\protect\citeauthoryear{Lewis}{1979}]{lewis}
D.~Lewis.
\newblock Score-keeping in a language game.
\newblock In {\em Semantics from Different Points of View}. Berlin: Springer,
  1979.

\bibitem[\protect\citeauthoryear{McGlashan}{1992}]{McGlashan}
Scott McGlashan.
\newblock {\em Towards a Cognitive Linguistic Approach to Language
  Comprehension}.
\newblock PhD thesis, University of Edinburgh, 1992.

\bibitem[\protect\citeauthoryear{Muskens}{1993}]{muskens:cdrt}
Reinhard Muskens.
\newblock A Compositional Discourse Representation Theory.
\newblock In {\em Proceedings of the Amsterdam Colloquium}, pages 467--468,
  1993.

\bibitem[\protect\citeauthoryear{Pustejovsky}{1991}]{pus:GL}
James Pustejovsky.
\newblock The Generative Lexicon.
\newblock {\em Computational Linguistics}, 17(4), 1991.

\bibitem[\protect\citeauthoryear{Pustejovsky}{1993}]{pus:TC}
James Pustejovsky.
\newblock Type Coercion and Lexical Selection.
\newblock In James Pustejovsky (ed.), {\em Semantics and the Lexicon}. Kluwer,
  1993.

\bibitem[\protect\citeauthoryear{Pustejovsky}{1995}]{pus:book}
James Pustejovsky.
\newblock {\em The Generative Lexicon: A theory of computational lexical
  semantics.}
\newblock MIT Press, Cambridge, MA, 1995.

\bibitem[\protect\citeauthoryear{Sanford and Garrod}{1981}]{SanfordandGarrod}
A.J. Sanford and S.C. Garrod.
\newblock {\em Understanding Written Language}.
\newblock Wiley \& sons, 1981.

\bibitem[\protect\citeauthoryear{Van~der Sandt}{1992}]{sandt:presupdrt}
Rob~A. Van~der Sandt.
\newblock Presupposition Projection as Anaphora Resolution.
\newblock {\em Journal of Semantics}, 9, 1992.

\end{thebibliography}
\end{document}